# Multi-views Fusion CNN for Left Ventricular Volumes Estimation on Cardiac MR Images

Gongning Luo[#], *Student Member, IEEE*, Suyu Dong[#], Kuanquan Wang*, *Senior Member, IEEE*, Wangmeng Zuo, *Senior Member, IEEE*, Shaodong Cao, and Henggui Zhang

*Abstract—Objective*: Left ventricular (LV) volumes estimation is a critical procedure for cardiac disease diagnosis. The objective of this paper is to address direct LV volumes prediction task. *Methods*: In this paper, we propose a direct volumes prediction method based on the end-to-end deep convolutional neural networks (CNN). We study the end-to-end LV volumes prediction method in items of the data preprocessing, networks structure, and multi-views fusion strategy. The main contributions of this paper are the following aspects. First, we propose a new data preprocessing method on cardiac magnetic resonance (CMR). Second, we propose a new networks structure for end-to-end LV volumes estimation. Third, we explore the representational capacity of different slices, and propose a fusion strategy to improve the prediction accuracy. *Results*: The evaluation results show that the proposed method outperforms other state-of-the-art LV volumes estimation methods on the open accessible benchmark datasets. The clinical indexes derived from the predicted volumes agree well with the ground truth (EDV: $R^2$=0.974, RMSE=9.6ml; ESV: $R^2$=0.976, RMSE=7.1ml; EF: $R^2$=0.828, RMSE =4.71%). *Conclusion*: Experimental results prove that the proposed method may be useful for LV volumes prediction task. *Significance*: The proposed method not only has application potential for cardiac diseases screening for large-scale CMR data, but also can be extended to other medical image research fields.

## I. INTRODUCTION

CARDIAC diseases are leading causes of death in the world [1]. The left ventricle (LV) is the biggest chamber in heart, which plays an important role in maintaining the ejection function of heart. The ejection fraction (EF) is a critical index for clinical diagnosis of cardiac diseases. Generally, the computation of left ventricular EF depends on the accurate estimation of LV volumes, which include end-diastole volumes (EDV) and end-systole volumes (ESV). The traditional volumes estimation methods on cardiac magnetic resonance (CMR) images are based on the LV segmentation technology. Besides, due to high discrimination of the soft tissue and the invasive exercise standards, CMR is considered as the gold standard modality for cardiac diseases diagnosis [2]. Hence, most of the cardiac medical image processing researches focus on CMR images, especially in the automatic LV segmentation tasks [3].

However, automatic LV segmentation is still an open and challenging task, due to the inherent characteristics of the cardiac MR images [4], such as higher noise, intensity level inhomogeneity, effect of partial volume [5, 6], complex topological structures, and great variability across different slices. Before 2011, the LV segmentation methods can be categorized into four kinds: 1) The methods based on traditional image-driven technologies, e.g., threshold based methods [5, 7], dynamic programming based methods [8], registration based methods [9], and graph based methods [10]. 2) The methods based on a deformable model, such as snake and level set [11-14]. 3) The methods based on pixel classification, e.g., cluster [15], neural networks [16] and Gaussian mixture model [17]. 4) The methods based on prior statistical information, for example, the active shape model (ASM) [18], the active appearance model (AAM) [19, 20] and the atlas model [21]. More detailed information about segmentation methods on CMR images before 2011 can be found in [3]. After 2011, some hybrid models have been proposed. For example, [22] proposed an LV segmentation method based on topological stable-state threshold and region restricted dynamic programming. [23] used Gaussian-mixture model and region restricted dynamic programming to segment LV in CMR images. Also, the deep-learning (DL) technology combined with deformable model was adopted to address LV segmentation problem [24]. What's more, some popular end-to-end semantic segmentation methods, such as fully convolutional neural networks (FCN) [25] and U-net [26], were also applied in varies of medical image segmentation tasks [27], which show the potential of the deep learning technology on clinical practice in some extent. Additionally, some new methods were also applied to LV segmentation field, for instance, the guided random walks method [28], the convex relaxed distribution matching method [29], and the mutual context information method [30]. Nevertheless, in the practical clinical diagnosis, most of the working LV segmentation methods are still semi-automatic, which are time-consuming

This work was supported by the National Key R&D Program of China under Grant No. 2017YFC0113000, and by the National Natural Science Foundation of China under Grant No. 61571165 and 61572152. (Corresponding author: Kuanquan Wang, email: wangkq@hit.edu.cn; # Joint first author)
G. Luo, S. Dong, K. Wang, and W. Zuo are with the School of Computer Science and Technology, Harbin Institute of Technology, Harbin 150001, China (e-mail: luogongning@hit.edu.cn; dongsuyu@hit.edu.cn; cswmzuo@gmail.com).
S. Cao is Department of Radiology, The Fourth Hospital of Harbin Medical University, Harbin 150001, China. (e-mail: shaodongcao@163.com).
H. Zhang is with School of Physics and Astronomy, University of Manchester, Manchester, UK (e-mail: H.Zhang-3@manchester.ac.uk).

and laborious for physicians to analyze large numbers of CMR images.

In recent years, some direct LV volumes estimation methods without segmentation have been attempted. For instance, [31] proposed an adapted Bayesian method to estimate the ventricle volumes, and this method had been validated on 56 subjects. [32] proposed a linear support vector machine method to assess LV myocardial function directly, and this method had been evaluated in 58 subjects. Furthermore, the method based on multi-scale deep belief networks and regression forests was proposed to direct ventricle volumes estimation, and this model had been trained and validated in 100 subjects [33]. These works show the research potential of direct volumes estimation in some extent. However, due to the lack of large labeled benchmark datasets, the generalization of these direct estimation methods is limited.

Fortunately, the 2016 Data Science Bowl Cardiac Challenge Data in Kaggle (DSBCCD) [34] provides large numbers of CMR data, as well as the consistent ground truth and are open public accessible for research. To my best knowledge, the DSBCCD benchmark contains the most large-scale CMR data up to now. It provides a new opportunity for LV estimation research in some extend. In this paper, inspired by DSBCCD and the high performance level of the deep convolutional neural networks (CNN) in the field of natural images classification [35-38] and medical image processing [27], we proposed an end-to-end framework to address LV volumes estimation problem. The main contributions are the following aspects:

1) We proposed a data preprocessing method to achieve data normalization and automatic LV detection.

2) We designed a robust deep convolutional neural network to achieve accurate LV volumes estimation.

3) We explored the representational capacity of different slices of CMR, and proposed an efficient multi-views fusion strategy based on slices with high representational capacity.

The remainder of this paper is organized as follows. The section II describes the details of the proposed framework, which includes the LV detection, the design of CNN architecture, and the multi-views fusion strategy. The section III presents the experimental results including prediction accuracy, robustness, computational efficiency, and the comparison with state-of-the-art LV volumes estimation methods. In the section IV, we discuss the results and characteristics of the proposed framework, and propose a feedback system to make the proposed method more viable in a clinical system. Finally, we conclude this work and its research potential in the future in the section V.

## II. METHOD

The framework of the proposed LV volumes prediction method is shown in Fig.1. Similar to the general deep learning tasks, this pipeline contains three parts: data preprocessing, model training and LV volumes prediction based on CNN, and final EF computation. To achieve clear description, we described the crucial contributions to this pipeline in the three aspects: data preprocessing method in the normalization and interest of region (ROI) detection, deep CNN design, and multi-views fusion strategy.

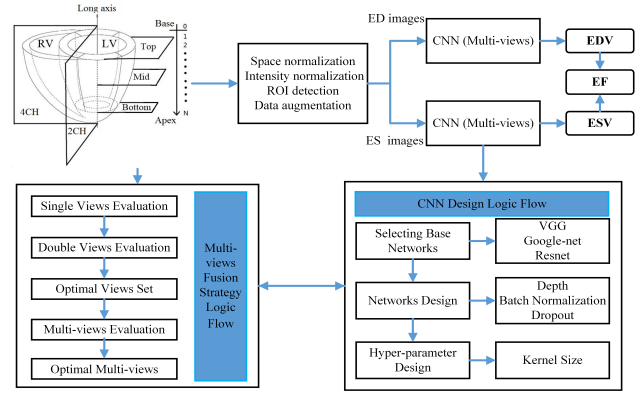

Fig.1. Illustration of the proposed frameworks of LV volumes prediction.

### A. Datasets and Metrics

#### 1) Datasets

In this paper, we used the dataset from DSBCCD. All the subjects in this dataset have public and consistent ground truth for EDV and ESV. According to the official DSBCCD benchmark, these data were split into 500 training subjects, 200 validation subjects, and 440 test subjects. The training and validation datasets were used for training and selecting optimal models in the section II. The test dataset was used for testing the performance of the selected model and comparing with the state-of-the-art methods in the section III. Additionally, we conducted the experiments based on python language and Keras [39] framework on Titan X GPU.

#### 2) Metrics

To study and evaluate the proposed framework, we adopted the root mean square errors (RMSE), mean of RMSE (MRMSE), and absolute errors (AE) as the metrics of the accuracy of volumes prediction. These metrics are defined as following:

$$RMSE = \sqrt{\sum_{i=1}^{n}(X_P - X_T)^2 / n} \quad (1)$$

$$MRMSE = (RMSE_{EDV} + RMSE_{ESV})/2 \quad (2)$$

$$AE = \sqrt{(X_P - X_T)^2} \quad (3)$$

where $X_P$ denotes the predicted value, $X_T$ is the ground truth value, $RMSE_{EDV}$ denotes the RMSE on EDV, $RMSE_{ESV}$ is the RMSE on ESV.

Specifically, in the section II, we adopted the RMSE and MERMSE to evaluate accuracy of the proposed methods on volumes prediction, and studied the effect factors of the framework design. In the section III, to achieve comprehensive evaluation of the proposed frameworks, we computed the RMSE and the standard deviation (SD) of AE (AESD) in 440 test subjects to evaluate the prediction results of the proposed method in terms of EDV, ESV, and EF. The EF value can be calculated in formula (4)

$$EF = (EDV - ESV)/EDV \quad (4)$$

What's more, in the section III, we also used regression analysis method and Bland-Altman (BA) analysis method to evaluate the correlation and agreement between the predicted values and the true values.

### B. Data Preprocessing

#### 1) Data normalization

To deal with variances existing on the adopted large-scale datasets, we proposed a normalization method to normalize the data in terms of the physical space and the intensity level. Firstly, we normalized the pixels' space of all the CMR images into 1.4 mm, which comes from the official recommendation of DSBCCD [34] based on the CMR imaging parameters statistics information of this dataset and the physical space is normalized using the formula (5):

$$WN = (PS/1.4)*W$$
$$HN = (PS/1.4)*H \qquad (5)$$

where *PS* refers to the actual physical space between adjacent pixels, *W* and *H* denote the width and height of CMR images respectively. *WN* and *HN* refer to the new width and height of resized images.

Generally, in a CMR image, the ventricle has higher intensity level, which is the most important indicator for volumes prediction. However, intensity level inhomogeneity exists in most of the data. Hence intensity level normalization must be done to enhance the contrast of CMR images. The intensity level normalization method is shown in (6):

$$X' = (X - Mean(X))/std(X) \qquad (6)$$

where $X$ refers to the input pixel intensity array, *std(X)* denotes standard deviation of $X$. *X'* is the final output value after intensity level normalization. In this way, we can normalize all the input data into uniform intensity level, which has high contrast.

#### 2) LV localization

As we all know, the region of interest (ROI) detection is critical for most of the computer vision tasks. Similarly, without a relative accurate ROI (LV region) detection method, the LV volumes estimation is impractical. Hence, we proposed a new LV detection method based on the intersections between long and short-axis image planes, and atlas mapping. First, we adopted the intersecting point to localize the coarse center point of LV, and then used the multi-scale atlas to achieve fine center point detection. As shown in the Fig.2, the 4 chambers (4CH) and 2 chambers (2CH) views in long-axis (LAX) and the views in short-axis (SAX) are intersecting in the 3D space. The intersecting points can be calculated easily, if the 4CH, 2CH and SAX coexist. We used the intersecting points as the coarse center points to crop 100*100 image patches, and then we proposed multi-scale atlas mapping method to find fine center points. Inspired by the traditional atlas generation method and CMR image registration method, the aim of the proposed atlas location method is to get the image patch which has the biggest similarity with the atlas. First, we selected 1000 SAX LV region patches (the size of every image is 64*64) through image cropping method. And then the mean image of the selected images was calculated, which is considered as an atlas. However, due to that the orientation and size of LV vary among different CMR images, we proposed multi-scale atlas method. That is to say, we extended the initial atlas into $R*T$ atlases through the resizing and rotation. *R* denotes the time of resizing operation (image uniformly-spaced resizing range from 52*52 to 72*72) and *T* denotes the time of rotation (image uniformly-spaced rotating range from 0 to 360 degrees).

To reduce the computation cost, we adopted the simple mean absolute deviation measure strategy to compute the similarity between an atlas and image patches as in

$$M = \frac{1}{R*T}\left(\sum_{c=1}^{R*T}\sum_{x=1}^{H_a}\sum_{y=1}^{W_a}|I_{x,y} - A_{x,y}^c|\right) \qquad (7)$$

where $H_a$ and $W_a$ denote the height and width of the current atlas respectively. $I$ and $A^c$ denote the current image patch and atlas respectively, *x* and *y* denote the coordinates of current pixel in $I$ and $A^c$. We slid the atlas in the target image to find the optimal matching patches with the minimal $M$, and cropped the 92*92 sizes image at the center of the patch as the final ROI. In the same way, LV localization in LAX views can also be detected. In practice, to further improve the calculation efficiency, we just detected the LV region in the top slice for short-axis, the ROI in mid and bottom slices can be gotten through direct ROI projection from the top slice. Based on the accurate LV detection method, we calculated the pixels' intensity sum in ROI for every frame, and selected frames corresponding to the maximum and minimum intensity sums as the ED and ES frames respectively. Besides, when some subjects lack the LAX slices, we also can directly adopt the atlas mapping method to find the LV location though such subjects are few in practice.

#### 3) Data augmentation

Finally, we augmented the data through rotating and shifting the input images during the training in every epoch. In this way, the robustness of the final prediction model can be enhanced. The details of implementation are shown as following:

**Input**: K training subjects.
**Output**: LV volumes Predicted model.
 1:  **For** *k*=1……1000 **do**
 2:      Data augmentation through image rotation;
 3:      Data augmentation through image shift;
 4:      Fitting model using "Adam" method;
 5:      Get prediction validation loss value of current model;
 6:      Get weights of current model;
 7:      **If** validation loss value is minimum **do**
             Save the weights of current model;
 8:      **End If**
 9:  **End For**

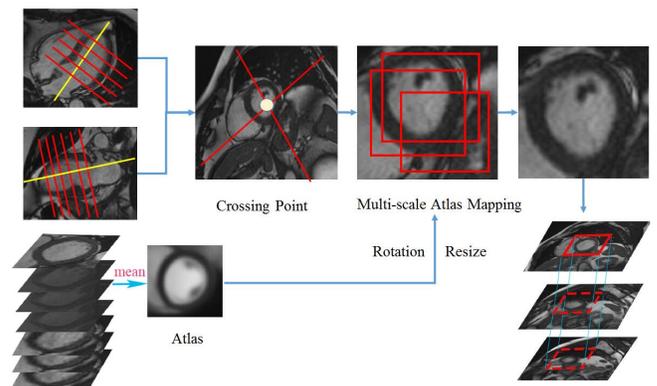

Fig.2. The LV detection method based on intersecting views and atlas mapping.

*C. CNN design*

With the development of deep learning technology, different kinds of network architectures were proposed for various applications. Actually, most of the improved deep learning architectures for the regression and classification tasks are based on the three basic networks: VGG-net [36], Google-net [37], and Residual-net [38]. Generally, the VGG-net follows the traditional CNN style, which consists of the convolutional layers, pooling layers, and fully connection (FC) layers sequentially. Google-net improves the traditional architecture through adding the multi-scale convolutional layers. Residual-net adopts an identity mapping method to make networks converge fast and avoid the gradient vanishing. In this work, we firstly evaluated the three basic networks in our task and proposed an improved method on the basic networks architecture.

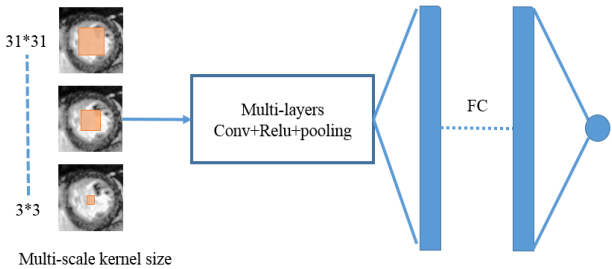

Fig.3. The CNN networks for LV volumes prediction.

In this paper, we formulated traditional classification networks into the end-to-end regression networks, and replace the softmax activation function of the last FC layers with rectified linear units $f(x)=\max(0,x)$ (Relu). As shown in Fig. 3, in the last layer, a fully connected layer containing one node was added for final output. We resized the input image into 224*224 and adopted the uniform weight to initialize the first convolutional layer and the last FC layer (because the number of input channels is different from the traditional RGB images, and the last FC layer does not exist in the natural images classification tasks), and the other layers were initialized by the pre-trained CNN model on ImageNet datasets [40] The loss function for optimization is the formula (8)

$$Loss = \sqrt{\sum_{i \in X}(X_i - Y_i)^2 / N} \qquad (8)$$

where $N$ refers to the number of training examples, $X_i$ denotes prediction value of the $i$th example, and $Y_i$ denotes the ground truth of the $i$th example. In this section, we trained the networks using Adam optimization method [41]. We selected the model with the minimum validation loss and evaluated the model based on RMSE and MRMSE. According to the official DSBCCD benchmark, in this section, we adopted 500 subjects for model training, and 200 subjects for model validation (we removed the cases whose LAX slices is missing, though such cases are few). The designed networks are not sensitive to most of the hyper parameters and have high robustness, however, to achieve repeatable experiment results, we fixed some hyper parameters to evaluate the proposed method. In this paper, the LV volumes prediction results of the proposed method are based on following setting. The "Adam" learning rate is 0.0001, the batch size is 64, and the number of epochs is 1000 for training EDV or ESV prediction model.

Besides, to find proper model, we conducted the two stages models evaluation. In the first stage we evaluated the three popular basic types of CNN architectures (the 19 layers VGG [36], Google-net [37], and the 50 layers Residual-net (Resnet 50) [38]) with the added single node layer in the end. In the second stage, we improved the design of the selected CNN architecture. We evaluated the prediction results based on the combination of different CNN models and views (the 4CH and 2CH views in LAX; the top, mid, and bottom views in SAX). Note that the details of views selection method will be shown in section D.

As we can see in the Table I, the VGG-net achieves the lowest RMSE with any input views. Hence, we adopted the VGG net as the basic network. Additionally, to achieve concise and clear presentation and show the obvious tendency of volumes prediction accuracy with the related factors, such as CNN depth, batch normalization, dropout and kernel size, we adopted the MRMSE as the basic metric in the following parts in this section.

As described in [36], the depth of CNN is crucial for most of deep learning tasks, in this paper, we firstly evaluated the influence of depth for prediction accuracy. As we can see in the Table III, the VGG with depth 20 achieves the lowest average RMSE. However, the computational cost increases with the increase of the depth of CNN. Besides, considered to the magnitude of accessible data, we did not try to improve the networks through adding more layers. Hence, we selected the VGG20 in Table II as the basic VGG. We improved the basic VGG in the following aspects:

*1) Batch normalization:* During the training, the distribution of features mapping in each layer changes acutely [42], because of the change of input from the previous layers. With the increasing of network layers, this phenomenon will be more apparent. Hence, the networks are very sensitive to initial parameters and learning rates. To enhance the robustness of the parameters, we added the batch normalization (BN) layers [42] following every convolutional layer.

*2) Dropout:* Additionally, in order to avoid over fitting problem in the designed networks, we adopted the "Dropout" method in [43] to drop out 25% hidden neurons after the third fully connected layers.

*3) Kernel size:* As we can see in the Table III, the proposed VGG20-BN architecture achieves the lowest RMSE. However, to further improve the performance of the proposed CNN, we made an assumption that the size of receptive field in the first convolutional layer is crucial for LV volumes estimation. As we can see in the Fig.3, the pink rectangle denotes the size of the receptive field in the first layer. Besides, we can make further assumption that the blood pool region of a CMR image is the most important indicator for volumes and the large receptive field can generate good features for prediction. To prove the proposed assumption, we tried to train and validate the CNN model with the various kernel sizes in the first layer based on VGG20-BN model in the Table III. As shown in the Fig. 4, with the increase of kernel size, the mean RMSE reduce obviously. This phenomenon proves the proposed assumption in some extend. Besides, we also find that too larger kernel size does not generate accuracy gain for prediction. Hence, according to the indication in the Fig. 4, values between 19 and 23 are suitable size of the receptive field in the first layer.

Considering the computation efficiency and number of parameters, we selected the 19 as the final kernel size in the first layer for the following experimental evaluation.

TABLE I
THE EVALUATION RESULTS OF DIFFERENT CNN MODEL WITH RMSE IN VALIDATION SET (P<0.001).

| Input views | VGG19+1 (ml) | | Google-net+1 (ml) | | Resnet50+1 (ml) | | AVG (ml) |
|---|---|---|---|---|---|---|---|
| | EDV | ESV | EDV | ESV | EDV | ESV | |
| 2CH | **13.1** | **10.1** | 16.3 | 13.5 | 17.5 | 13.6 | **14** |
| 4CH | 14.5 | 11.9 | 17.1 | 14.1 | 18.1 | 14.1 | 15 |
| 2CH+4CH | 14.3 | 12.1 | 16.3 | 13.9 | 17.3 | 14.2 | 14.7 |
| Top | 13.2 | **9.6** | 17.2 | 12.9 | 17.2 | 13.9 | **14** |
| Mid | 13.5 | 10.6 | **16.6** | **13.5** | **17.6** | **14.1** | **14.3** |
| bottom | 14.1 | 12.5 | 19.8 | 17.9 | 18.1 | 14.9 | 16.2 |
| T+M+B | 13.7 | 11.3 | 16.9 | 12.3 | 17.8 | 13.9 | **14.3** |
| Apex | 15.1 | 13.3 | 21.1 | 16.9 | 23.1 | 15.7 | 17.5 |
| Base | 15.7 | 13.9 | 20.7 | 17.8 | 21.5 | 15.1 | 17.5 |
| AVG(ml) | 14.1 | 11.7 | 18 | 14.8 | 18.7 | 14.4 | |

This table shows the mean RMSE values of EDV and ESV in various CNN architectures with different input views. The best performances in each case are shown in bold face. The AVG denotes the average value and the T+M+B denotes Top+Mid+Bottom input views combination. Note that all the three CNN architectures are added one layer, which only includes one node in the last layer (Actually, the VGG 19 is same with the VGG 20 in Table II, in which the FC-1 is added in the last layer).

TABLE II
THE NETWORKS ARCHITECTURE.

| VGG14 | VGG17 | VGG20 | VGG20-BN |
|---|---|---|---|
| Input(224*244) | | | |
| Conv3-64*2 | Conv3-64*2 | Conv3-64*2 | Conv3-64*2-BN |
| Maxpooling layer (kernel size=2 stride=2) | | | |
| Conv3-128*2 | Conv3-128*2 | Conv3-128*2 | Conv3-128*2-BN |
| Maxpooling layer (kernel size=2 stride=2) | | | |
| Conv3-256*2 | Conv3-256*3 | Conv3-256*4 | Conv3-256*4-BN |
| Maxpooling layer (kernel size=2 stride=2) | | | |
| Conv3-512*2 | Conv3-512*3 | Conv3-512*4 | Conv3-512*4-BN |
| Maxpooling layer (kernel size=2 stride=2) | | | |
| Conv3-512*2 | Conv3-512*3 | Conv3-512*4 | Conv3-512*4-BN |
| Maxpooling layer (kernel size=2 stride=2) | | | |
| FC-4096 | | | |
| FC-4096 | | | |
| FC-1000 | | | |
| --- | | | dropout |
| FC-1 | | | |

The convolutional module parameters are denoted as Conv (receptive field size) - (number of channels * number of layers). The fully connected module parameters are denoted as FC - (number of node). The depths of networks increase from 14 to 20 from left to right.

### D. Multi-views fusion strategy

Inspired by the multi-channels fusion method in natural image processing field [44], the multi-views fusion strategy, is widely used in medical image classification problems [45]. Generally, the multi-views fusion strategy on the first layer of CNN is based on the assumption that more effective features can be extracted through multi-views input. However, in some cases, inappropriate views input can lead to decrease of the prediction accuracy. In this paper, as shown in the Table I and III, the best results do not come from the multi-views fusion, such as the fusion of top, mid, and bottom slices in SAX or the fusion of 2CH and 4CH in LAX. Hence, to find the best combination strategy, we proposed a views combination optimal set analysis strategy. We conducted the experiments in the two stages. In the first stage, we evaluated the performance on any input combination of double views. In the second stage, the views in top 4 view-pairs were selected to construct the optimal views set, and we evaluated the performance on views input combinations from optimal views set.

TABLE III
THE EVALUATION RESULTS OF DIFFERENT CNN MODELS IN VALIDATION SET (P<0.001).

| Input Views | VGG 14 (ml) | VGG 17(ml) | VGG 20 (ml) | VGG 20-BN (ml) | AVG (ml) |
|---|---|---|---|---|---|
| 2CH | 11.6 | 11.3 | 11.2 | 9.8 | **10.9** |
| 4CH | 13.2 | 12.9 | 11.6 | 10.4 | 12 |
| 2CH+4CH | 13.2 | 13.7 | 13.9 | 11.9 | 13.2 |
| Top | 11.4 | 11.2 | 11.2 | 10.7 | 11 |
| Mid | 12.05 | 11.9 | 11.5 | 10.9 | 11.6 |
| bottom | 13.3 | 13.5 | 12.3 | 11.9 | 12.8 |
| T+M+B | 12.5 | 11.3 | 10.5 | 11.6 | 11.5 |
| Apex | 14.2 | 14.1 | 14.1 | 13.5 | 14 |
| Base | 14.8 | 14.3 | 14.1 | 13.9 | 14.3 |
| AVG(ml) | 12.9 | 12.7 | 12.3 | **11.6** | |

This table shows the MRMSE values in different VGG networks architecture. The best performances in each case are shown in bold face. The AVG denotes the average value and the T+M+B denotes Top+Mid+Bottom input views combination.

First, to achieve consistent expression, we define the views selection method. Usually, we classify slices into two kinds: the slices from SAX and slices from LAX. In the SAX, the amount and thicknesses of slices are various among different subjects because of the different settings of MRI scanning parameters. But in the LAX there are only the 4CH slice and the 2CH slice. The reference topological structures of LV in different slices are shown in Fig.5. In this paper, we define the first slice in SAX as the base view, the last slice in the SAX as the apex view, the second slice as the top view, the second last slice as the bottom view and the other slices as the mid views.

Due to the variation of the amount of slices on SAX from top to bottom (usually, the number of slices varies range from 8 to 20), the selection of the mid slice in SAX is necessary for uniform data input. We design a sampling method as in:

$$M = [1 + C/2] \qquad (9)$$

where $C$ refers to the amount of slices from top to bottom in SAX, M denotes the number of mid slice that we want to select.

As is shown in Table I and III, when we used the apex and base views as inputs, all the CNN models have the highest RMSE. Hence, we can make a conclusion that the base and apex slices have lower representational capacity for CNN than other views because of complex topological structures in the SAX base slice and the unapparent ventricle chamber in the SAX apex slice. Based on this conclusion, the following combination matrix analysis does not adopt the apex and base views as input. As shown in the Table IV, we evaluated the any input combination of double views among the 2CH, 4CH, top, mid, and bottom views on the mean RMSE metrics. And then the top 4 view-pairs were found: <2CH, 4CH> <2CH, Top> <2CH, Mid> and <Top, Mid>. Besides, the views in top 4 view-pairs were selected to construct the optimal views set:

{2CH, 4CH, Top, Mid}. To further explore performance of the optimal views combination inputs, we evaluated views input combinations from optimal views set. As shown in the Table V, the combination of <Top, Mid, 2CH> achieves the lowest RMSE. Hence, the <Top, Mid, 2CH> was selected as the final multi-views fusion strategy.

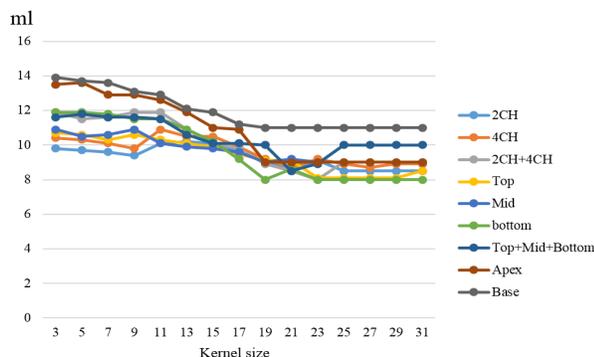

Fig.4. The MRMSE values in different kernel size of the first convolutional layers.

TABLE IV
THE EVALUATION RESULTS OF DOUBLE VIEWS IN VALIDATION SET ($P<0.001$).

| Views | 2CH | 4CH | Top | Mid | Bottom |
|---|---|---|---|---|---|
| 2CH | 9.1 | **8.7** | **8.2** | 8.6 | 9.1 |
| 4CH | - | 9.2 | 8.9 | 9.3 | 10.1 |
| Top | - | - | 9.1 | **8.3** | 10.5 |
| Mid | - | - | - | 9.2 | 9.8 |
| Bottom | - | - | - | - | 9.9 |

This table shows the MRMSE values in double views combination input. The best performances in each case are shown in bold face.

TABLE V
THE EVALUATION RESULTS OF MULTI-VIEWS FUSION IN VALIDATION SET ($P<0.001$).

| Views | EDV | ESV |
|---|---|---|
| Top+Mid+2CH | **8.1** | **6.9** |
| Top+Mid+4CH | 9.2 | 8.1 |
| 4CH+2CH+Top | 9.3 | 8.5 |
| 4CH+2CH+Mid | 9.1 | 8.2 |
| Top+Mid+4CH+2CH | 9.3 | 7.9 |

This table shows the RMSE values of EDV and ESV in optimal views combination. The best performances in each case are shown in bold face.

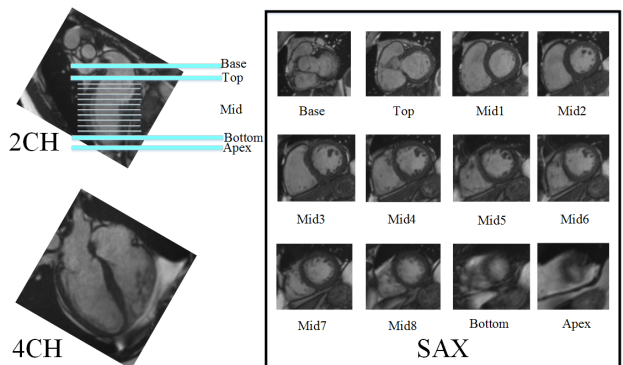

Fig.5. The slices from base to apex in short axis, 4CH and 2CH slices in long axis.

## III. EXPERIMENTS

In this section, we showed some experiment results to demonstrate that the proposed LV volumes prediction method is reasonable and can achieve accurate LV volumes estimation.

### A. Setting

In this section, based on the validation results in section II, we selected the 2CH, Top and Mid slices as the input views, the input images were resized into 224*224, hence the final input dimensionality is 3*224*224. However, some subjects lost the 2CH slices, even though in practice these subjects are few. To achieve complete evaluation on the 440 test subjects, we also prepared the backup models, whose input data are only top and mid views from SAX. Additionally, we trained the proposed CNN three times to generate three different prediction models on the same CNN architecture, and the final prediction results are the mean of three prediction results from the three models.

### B. Volumes Prediction Results

In this subsection, we reported the volumes prediction results in terms of ESV, EDV, and EF. On the whole, the proposed method achieves high accuracy in 440 test subjects. The RMSE±AESD in ESV, EDV, and EF are 7.1±3.36 (ml), 9.6±4.9 (ml), and 0.047±0.018 respectively ($P<0.001$).

The correlation graphs in Fig.6 (a), (c), and (e) show the correlation between the ground truth and predicted results. The correlation coefficients (R) achieve 0.976, 0.973, and 0.828 respectively. Additionally, the BA graphs in Fig.6 (b), (d), and (f) show the distribution of differences between the ground truth values and predicted values along the means between the ground truth values and predicted values. The means of differences achieve -2.5ml (with confidence intervals between -14.1 and 9.1), -6.6ml (with confidence intervals between -25 and 11.8), and 0 (with confidence intervals between -0.09 and 0.08) in terms of ESV, EDV, and EF, respectively.

### C. Comparison

We compared the prediction results with the top 4 teams' prediction results which come from the official statistic of DSBCCD, in terms of RMSE, correlation and agreement. As shown in the Table VI, the proposed method achieves the lowest error in the volumes RMSE and second place (4.71) in EF RMSE, which is little difference with the first place. Besides, in terms of the correlation and agreement, the proposed method outperforms the top 4 teams in estimation agreement of volumes and EF, and meanwhile achieves highest correlation. Actually, the FCN [25] and U-net [26] were used for LV volumes estimation in the Top 1 and Top 3 teams in the Table VI respectively. Hence, the Table VI also shows the superiority of the proposed method compared to the FCN and U-net, which are widely used for some end-to-end medical image segmentation tasks.

We selected two state-of-the-art methods to conduct comparison experiments, an indirect volumes estimation method through LV segmentation, which is based on DL and deformable model [24] and a direct volumes prediction method based on multi-scale filter and DL [33]. These two methods are trained on 700 subjects and tested on 440 test subjects in DSBCCD. To achieve the robustness analysis across different ages, we compared the proposed method with the two state-of-the-art methods in terms of the accuracy in ESV, EDV, and EF on different age groups. Considered to that the LV volumes vary among different age groups, we analyzed the average accuracy statistic results of three LV volumes estimation methods on every age group in Table VII. We can see that the proposed method achieves the lowest RMSE in the most of age groups. However, the other two methods obviously have higher RMSE than the proposed method. Besides, to my best knowledge, the previous algorithms are limited in the magnitude of datasets and have lower generalization. To some extent, the proposed method has higher robustness in big-scale datasets than the previous algorithms.

### D. Performance of LV Localization

The LV localization is crucial for direct LV volumes estimation based on deep learning. Hough transform [17] and temporal Fourier analysis [46] are widely used in LV localization task on CMR images processing. Hence, to evaluate the performance of the proposed LV localization method, we conducted experiments using the proposed method with replaced ROI detection methods using Hough transform (PMH) and temporal Fourier analysis (PMF). As shown in the Table VI, compared with PMH and PMF, the proposed method achieves best results, which can prove that the proposed LV localization method has better performance than the LV localization methods in [17] and [46] on our application.

### E. Robustness Evaluation

To evaluate the robustness of the proposed method, we adopted the models trained on the training datasets of DSBCCD to predict the LV volumes of the collected CMR data in the same way with above experiments. The collected CMR data all include infarct on a section of cardiac muscle in an area of the LV volume, which is a common case in the clinical application. We collected the 5 patients' CMR data. And we acquired 20~30 phases for every patient along the whole cardiac cycle. The phase's total of the five patients is 130. The ground truth of LV volume on every phase is the mean of volumes calculated through the manual delineation of LV endocardium by five physicians. The RMSE of 130 phases is 7.3 (P<0.001), which proves that the proposed method has high robustness on the clinical application.

### F. Efficiency

To evaluate the efficiency of the proposed methods, we analyzed the average processing time of LV volumes estimation on the proposed method and the state-of-the-art methods. The average processing times of one patient using the proposed method, the deformable model method (DLDM) [24] and the Multi-scale Filter Learning method (MFL) [33] are 100ms, 5000ms, and 1000ms respectively. Compared to the state-of-the-art LV volumes estimation methods, the proposed method markedly reduces the calculation time of LV volumes estimation on CMR. The high volumes estimation efficiency also shows the application potential of the proposed method on large-scale CMR data, which is important along with the increasing CMR data on the clinical application.

TABLE VI
THE COMPARISON RESULTS WITH TOP 4 TEAMS IN DSBCCD ON TEST SET (P<0.001).

| Method | RMSE | | | AV (ml) | AEF (%) | RV (ml) | REF (%) |
|---|---|---|---|---|---|---|---|
| | EDV (ml) | ESV (ml) | EF (%) | | | | |
| PM | **9.6** | **7.1** | 4.71 | **18** | 8.1 | 0.97 | **0.82** |
| PMH | 11.9 | 9.9 | 5.38 | 21.5 | 9.7 | 0.97 | 0.79 |
| PMF | 11.5 | 9.6. | 5.12 | 21.6 | 9.2 | 0.97 | 0.81 |
| Top1 | 12.02 | 10.19 | 4.88 | 21.8 | 9.1 | 0.97 | 0.78 |
| Top2 | 13.65 | 10.43 | 6.99 | 23.8 | 8.8 | 0.97 | 0.78 |
| Top3 | 13.63 | 10.32 | 5.04 | 23.7 | 9.6 | 0.97 | 0.76 |
| Top4 | 13.2 | 9.31 | **4.69** | 22.4 | 9.2 | 0.97 | 0.77 |

This table shows the comparison results between the proposed method (PM) and other methods (including the proposed method with replaced ROI detection methods using Hough transform [17] and temporal Fourier analysis [46], and top 4 teams from DSBCCD). PMH and PMF denote the proposed framework with different LV localization methods using Hough transform and temporal Fourier analysis respectively. AV and AEF denote the agreement (1.96* SD of difference between predicted and true values) with ground truth in terms of volumes and EF respectively. RV and REF denote the correlation with ground truth in terms of volumes and EF respectively. The best performances in each case are shown in bold face.

TABLE VII
THE COMPARISON RESULTS WITH TWO STATE-OF-THE-ART METHODS ON TEST SET (P<0.001).

| Age (number) | ESV(ml) | | | EDV(ml) | | | EF (%) | | |
|---|---|---|---|---|---|---|---|---|---|
| | PM | DLDM | MFL | PM | DLDM | MFL | PM | DLDM | MFL |
| <10(23) | 3.9(**2.3**) | 4.9(3.3) | **3.8**(5.9) | **6.1(6.4)** | 7.1(11.3) | 12.5(11.1) | **7.9(13.1)** | 15.9(19.1) | 17.9(15.1) |
| 10-20(79) | **3.7(4.4)** | 9.7(4.9) | 8.1(7.2) | **7.2(5.5)** | 15.9(16.8) | 19.7(9.8) | **5.3(3.2)** | 9.7(4.9) | 8.7(7.3) |
| 20-30(44) | **5.3(3.3)** | 11.5(9.3) | 9.5(13.8) | **8.1(6.7)** | 21.9(15.2) | 21.3(19.3) | **3.4(2.1)** | 11.3(13.2) | 9.4(8.7) |
| 30-40(41) | **6.6(3.5)** | 10.1(13.5) | 7.3(5.2) | **8.5(5.2)** | 13.9(20.1) | 28.4(13.1) | **3.0(2.3)** | 8.1(8.5) | 7.6(9.9) |
| 40-50(63) | **8.8(3.8)** | 9.4(5.8) | 9.8(7.9) | **9.7(8.9)** | 19.7(15.8) | 14.8(23.3) | **3.3(3.2)** | 11.8(9.8) | 12.9(5.8) |
| 50-60(81) | **5.2(3.1)** | 12.2(3.9) | 8.3(5.1) | **10.3(5.3)** | 23.6(6.9) | 24.3(23.1) | **4.8(2.1)** | 10.1(13.6) | 15.2(14.1) |
| 60-70(70) | **7.6(3.3)** | 9.1(10.3) | 13.6(11.3) | **9.8(5.6)** | 15.1(15.8) | 25.4(21.9) | **3.7(1.2)** | 9.6(8.3) | 8.6(11.3) |
| >70(39) | **7.6(2.5)** | 17.7(11.3) | 7.8(9.9) | **11.3(8.4)** | 21.1(11.4) | 25.6(20.5) | **3.7(3.3)** | 8.1(8.2) | 10.1(6.9) |

This table shows the comparison results between the proposed method (PM) and two state-of-the-art methods, and those are DL with deformable model method (DLDM) [24] and Multi-scale Filter Learning method (MFL) [33]. The numbers format is RMSE value (AESD). The best performances in each case are shown in bold face.

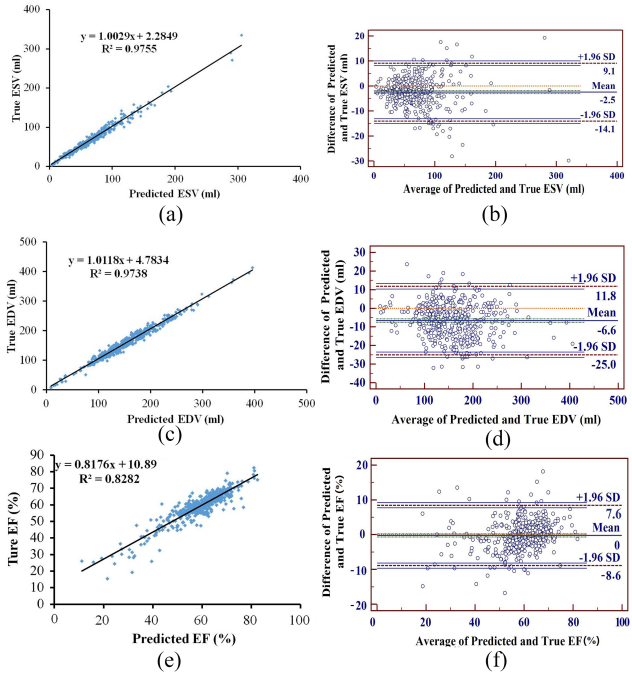

Fig.6. Correlation graphs and Bland-Altman graphs (P<0.001). (a)The correlation between the predicted ESV and the true ESV. (b) The Bland-Altman analysis between the predicted ESV and the true ESV. (c)The correlation between the predicted EDV and the true EDV. (d) The Bland-Altman analysis between the predicted EDV and the true EDV. (e)The correlation between the predicted EF and the true EF. (f) The Bland-Altman analysis between the predicted EF and the true EF.

## IV. DISCUSSION

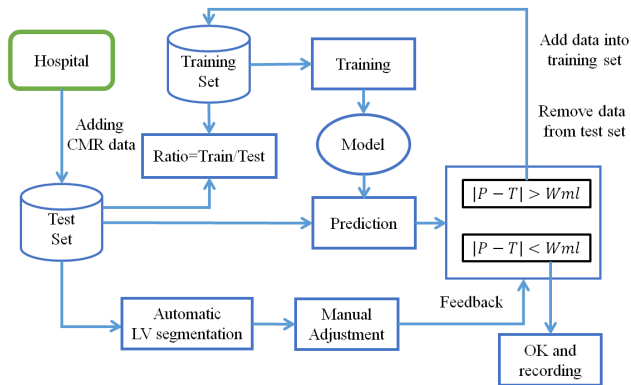

Fig.7. The flowchart of the proposed feedback system. In this picture, P denotes the prediction values from deep learning model, T denotes the ground truth values from manual label of clinical experts, W denotes the threshold of acceptable absolute error between T and P.

In this study, we developed a direct LV volumes estimation method based on end-to-end DL technology. The critical improvement is based on the proposed LV detection method, CNN architecture and multi-views fusion strategy. Compared with traditional LV segmentation methods, we converted the LV volumes estimation problem to the end-to-end regression problem avoiding some difficulties existing in CMR images segmentation. Also, the proposed framework improves LV volumes prediction in terms of the prediction accuracy, robustness, efficiency, and generalization ability. The high LV volumes prediction accuracy in large-scale dataset also proves these points. In this section, the discussion is shown on the following three aspects: characteristics of the proposed method, feedback system, and limitation.

### A. Characteristics of the proposed method

In the aspect of data preprocessing, we proposed an accurate and robust ROI detection method. Besides, a normalization method to handle the variance in pixel space and intensity level for large number of CMR dataset was proposed. The preprocessing step not only improves the prediction accuracy, but also improves the proposed framework's generalization ability and robustness in large-scale data, which come from various vendors and scanning parameters, and consist of large number of subjects with different ages, genders and health levels.

In the aspect of DL networks design, we evaluated the popular convolutional networks architecture and improved the VGG adopting some effective technologies, such as batch normalization, "Adam" training, and dropout. Besides, relationship between the prediction accuracy and the size of receptive field in the first convolutional layer was found, this finding enable us further improve the design of CNN.

In the aspect of multi-views fusion, we proposed a multi-views fusion strategy based on the views combination optimal set analysis strategy. Through lots of experiments, we studied the prediction ability of single slice, and used the slices with high representational ability (the optimal multi-views combination) as input for CNN to further improve the prediction accuracy. What's more, the proposed framework based on multi-views has superiority on the cases with slices missing. Because the single-views and double-views models can also be used as backup for LV volume prediction.

Compared to the volumes prediction methods based on the segmentation, the obvious superiority is that the proposed method is robust to the variances of slices space due to the different scanning parameters of CMR equipment. However, the traditional estimation methods are sensitive to the cases with large space between adjacent slices or some cases of slices missing, because some LV volumes estimation methods based on segmentation generally calculate the volumes through the accumulation of LV region segmented in every slice.

### B. Feedback system

Though the proposed method achieves high prediction performance on DSBCCD, the drawback of direct estimation methods in comparison to segmentation approaches is that the results' feedback to clinical experts is difficult. As we all know, the deep learning has achieved good performance and high prediction efficiency on a lot of tasks, such as papers [25], [26], and [27]. However, the deep learning applications face the problems of feedback and interpretability. Nevertheless, feedback is an important part on more clinically viable system, hence a prediction system with good feedback is necessary. Fortunately, the advantage of deep learning application is that the more training data are acquired and the higher performance will be achieved. Some successful deep learning applications have proved that the deep learning models will outperform human experts if they are trained on enough large datasets [27]. The good performance on DSBCCD also shows the potential of the proposed method on achieving higher volumes prediction accuracy along with the increase of large-scale clinical CMR data. To achieve wider application on clinically viable system, we proposed a feedback system through comparing with the ground truth from automatic or semi-automatic LV segmentation methods (such as [5, 6], [7] and [10]) combining the human experts' adjustment. As shown in the Fig.7, the new CMR data from hospitals can be added into the test datasets. These data can be labeled by some automatic or semi-automatic LV segmentation methods, and then the true LV volumes can be gotten by some experts' further adjustment and calculation. Ideally, the mean values of LV volumes from more than 3 experts are considered as the ground truth values for feedback validation of the proposed LV volumes estimation method. If the absolute value of the difference between the ground truth value ($T$) and the prediction value ($P$) is larger than the threshold ($W$), such test data will be removed from test set and added into training set. Otherwise, we consider the feedback is good and it will be recorded. Finally, with the increasing of the training datasets and the iterative training of deep learning model (We can retrain the model when some new CMR data are added into the training set), the proposed feedback system will converge, when the ratio between the number of training set and test set is lower than $R$ (We can assume $R$=0.1) and the continuous $F$ good feedbacks are recorded (We can assume $F$=1000). In this way the generalization ability of the model can be guaranteed. Note that the $R$ and $F$ can be adjusted in practical clinical application.

The proposed feedback system can not only get feedback but also further improve the performance of prediction. Finally, with the increase of the CMR data, the prediction accuracy of the deep learning model may approach the accuracy of manual label. Though it is a hard and time-consuming work to validate the proposed feedback system because it needs the support of the enough labeled CMR data, we believe that the enough labeled CMR data will be available in the future along with the popularization of deep learning application in clinical practice. When the feedback system converges, the trained deep learning model can be used in a viable application, such as cardiac diseases screening on large-scale CMR data.

*C. Limitation*

Additionally, though the proposed method achieves high prediction accuracy and outperforms state-of-the-art methods on the adopted datasets, we also find some phenomena on final testing results and limitation of the adopted datasets, which may be useful for further improvement research on this topic.

Firstly, EDV and ESV is critical index for clinical diseases diagnosis, such as the ventricular dilatation or remodeling [5]. Hence, accurate estimation of EDV and ESV is meaningful inherently. In theory the high volumes estimation accuracy denotes the high EF estimation accuracy. However, in fact, the randomness of difference between the predicted values and ground truth values results in the uncontrolled EF prediction error. For example, we assume the true EDV=140ml and the true ESV=60ml, hence, the true EF = (140-60)/140=57.1%. We further assume that the RMSE of EDV is 10ml, and RMSE of EDV is 7ml. If the predicted EDV=140+10ml and the predicted ESV=60+7ml, the predicted EF = (150-67) / 150 = 55.3% (The difference with ground truth is 1.8%). If the predicted EDV=140-10ml and the predicted ESV=60+7ml, the predicted EF=(130-67)/130=48.5% (The difference with ground truth is 8.6%). If the predicted EDV=140+10ml and the predicted ESV=60-7ml, the predicted EF=(150-53)/150=64.7% (The difference with ground truth is 7.6%). If the predicted EDV=140-10ml and the predicted ESV=60-7ml, the predicted EF=(130-53)/130=59.2% (The difference with ground truth is 2.1%). We also can see in the Table VI that the Top 1 team achieves low RMSE in ESV and EDV, however, the Top 4 team achieves the best EF prediction results. Hence, we can find that the randomness of difference between the predicted values and ground truth values results in the uncontrolled EF prediction error, and the only way to improve the EF prediction accuracy is to further improve the volumes prediction accuracy.

Secondly, as we can see in the evaluation results, we found that accuracy in ESV is higher than in EDV overall.

Thirdly, as shown in Table VII, the predicted EF values have higher RMSE in the subjects whose ages are lower than 10. The imbalanced datasets distribution on training and test sets for every age group may result in the age-effect on the final predicted results, especially for the <10 age group (only 6% in the whole datasets). Besides, generally the LV volumes of nonage persons are obviously smaller than the LV volumes of adults. Hence, more CMR datasets of nonage persons are necessary for training more robust model. The age-effect will be further researched in the future if enough large datasets for every age group are gotten.

Additionally, prior works [5, 6] have demonstrated that the traditional method of delineating ground truth segmentations and measurements for cardiac volumes on CMR data using contours leads to significant error in volumetric estimation because of the existence of partial volume voxels. This also may be the possible flaws of DSBCCD in representing accurate measurements.

Finally, even though the proposed method improves performance on DSBCCD dataset, some potential limitations exist. The proposed multi-view fusion strategy throws away some slices during the volume prediction processing. In the future, some work about how to fuse all the CMR slices on LV region to improve the accuracy and generalization will be studied. Besides, though we proposed a feedback system, we must state that the proposed method provides no interpretability to the physician. In fact, the real interpretability on LV volumes prediction means that doctors must see which pixels were included in blood volume computations. In the future, the research about how to achieve interpretability on the direct LV volumes prediction task is a meaningful work.

These facts suggest us to study the relationship between EDV, ESV, EF, age and data flaws to achieve higher prediction accuracy cooperatively in the future. Besides, to further extend the application of the proposed framework, we will also study the other ventricular function indexes' prediction tasks, such as the ventricular mass, disease prediction, shape analysis and so on.

## V. CONCLUSION

In this paper, we proposed an end-to-end LV volumes prediction framework based on CNN to achieve accurate LV volumes prediction. The critical factors for the proposed framework are that the proposed ROI detection method, the designed CNN architecture, and the multi-views fusion strategy. The experiment results prove that the proposed method outperforms state-of-the-art LV volumes estimation method on the adopted datasets and has big potentials to be extended to other medical image research fields.

## ACKNOWLEDGMENT

Thanks for annual data science bowl in 2016, which provided the substantial for our research works. Thanks for Shaodong Cao, Chao Gao, Kezheng Wang, Weizhuo Hu and Dandan Dong, who provided the ground truth of LV volumes through manual delineation of LV endocardium. And we also thanks for the donations of Nvidia for Titan X. Additionally, the code is open accessible:

https://github.com/luogongning/Multi-views-fusion.

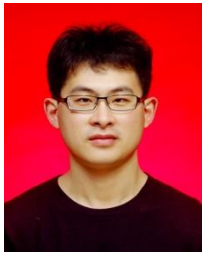
Gongning Luo received the B.Sc. degree in software engineering in East China Jiao Tong University and the M.Sc. degree in computer science from Harbin Institute of Technology in China, in 2012 and 2014, respectively. He is currently the Ph.D. candidate from the School of Computer Science and Technology, Harbin Institute of Technology, Harbin, China. He has published over 10 papers, and won the champion of blood vessel vulnerable plaque recognition challenge on CCCV 2017.

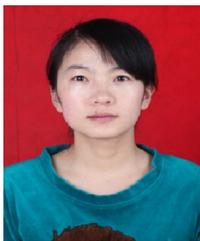
Suyu Dong received the B.Sc. degree in software engineering in East China Jiao Tong University and the M.Sc. degree in management science and engineering from Inner Mongol University of Technology in China, in 2012 and 2015, respectively. He is currently pursuing the Ph.D. degree from the School of Computer Science and Technology, Harbin Institute of Technology, Harbin, China. His research interests include deep learning, medical image processing and ultrasonic imaging.

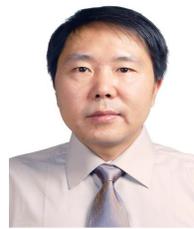
Kuanquan Wang (M'01–SM'07) is currently a Full Professor and Ph.D. Supervisor with the School of Computer Science and Technology, Harbin Institute of Technology. He is a Senior Member of the China Computer Federation and the Chinese Society of Biomedical Engineering. His main research areas include image processing and pattern recognition, biometrics, biocomputing, virtual reality, and visualization. He has published over 300 papers and 6 books, got more than 10 patents, and won a second prize of National Teaching Achievement.

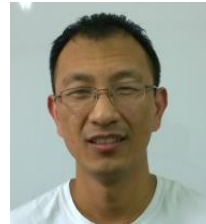
Wangmeng Zuo (M'09–SM'15) received the Ph.D. degree in computer application technology from the Harbin Institute of Technology, Harbin, China, in 2007. From 2004 to 2004, from 2005 to 2006, and from 2007 to 2008, he was a Research Assistant with the Department of Computing, Hong Kong Polytechnic University, Hong Kong. From 2009 to 2010, he was a Visiting Professor with Microsoft Research Asia. He is currently a Full Professor with the School of Computer Science and Technology, Harbin Institute of Technology. He has published more than 100 papers in top tier academic journals and conferences. His current research interests include image modeling and blind restoration, discriminative learning, biometrics, and 3D vision. He is an Associate Editor of the IET Biometrics

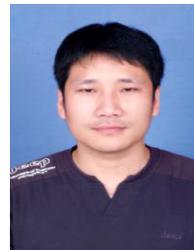
Shaodong Cao received the B.Sc. degree and the M.Sc. degree in Medical imaging from Harbin Medical University, Harbin, China, in 2007 and 2010, respectively. He is the senior chief engineer in the Fourth Hospital of Harbin Medical University, and he is also the director of Committee on imaging technology of Heilongjiang medical association. His research interests include CT and MR Imaging.

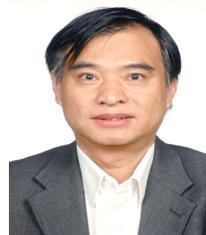
Dr. Henggui Zhang is Professor of Biological Physics. He received his Ph.D. degree in Mathematical Cardiology from the University of Leeds in 1994. Then he worked as postdoctoral research fellow at Johns Hopkins University School of Medicine (1994-1995) and the University of Leeds(1996-2000), and then senior research fellow at the University of Leeds(2000-2001). In October 2001, he moved to UMIST to take up a lectureship. From then, he worked as lecturer (2001-2004; UMIST), senior lecturer (2004-2006) and Reader (2006-2009) in the University of Manchester. He currently holds Chair of Biological Physics Group at School of Physics & Astronomy, The University of Manchester.